\documentclass{article}

\usepackage{arxiv}

\usepackage[utf8]{inputenc} 
\usepackage[T1]{fontenc}    
\usepackage{hyperref}       
\usepackage{url}            
\usepackage{booktabs}       
\usepackage{amsfonts}       
\usepackage{nicefrac}       
\usepackage{microtype}      
\usepackage{amsmath}
\usepackage{cleveref}       
\usepackage{lipsum}         
\usepackage{graphicx}
\usepackage{natbib}
\usepackage{doi}
\usepackage{algorithmic}
\usepackage{algorithm}
\usepackage{multirow}
\usepackage{makecell}
\usepackage{array}
\usepackage{wrapfig}

\title{\textbf{Adapt-Pruner:} Adaptive Structural Pruning for Efficient Small Language Model Training}

\date{}

\newif\ifuniqueAffiliation
\uniqueAffiliationtrue

\ifuniqueAffiliation 
\author{
Rui Pan$^1$\thanks{Equal contribution.} \quad \bf Shivanshu Shekhar$^1$\footnotemark[1]  \quad \bf Boyao Wang$^1$ \footnotemark[1] 
 \quad Shizhe Diao$^2$ \quad \bf Jipeng Zhang$^1$  \\ \bf Xingyuan Pan$^2$ \quad\bf Renjie Pi$^2$ \quad \bf Tong Zhang$^1$ \\
{$^1$ University of Illinois Urbana-Champaign} \quad
{$^2$ HKUST} \\
{\tt\small \{ruip4, shekhar6, boyaow2, xp12\}@illinois.edu, \quad \{sdiaoaa, rpi, jzhanggr\}@ust.hk} 
\\ {\tt\small tongzhang@tongzhang-ml.org} \\
}
\fi

\renewcommand{\shorttitle}{Adapt-Pruner: Adaptive Structural Pruning for Efficient Small Language Model Training}

\hypersetup{
pdftitle={Adapt-Pruner: Adaptive Structural Pruning for Efficient Small Language Model Training},
pdfauthor={Boyao Wang, Rui Pan},
}

\begin{document}
\maketitle

\begin{abstract}
Small language models (SLMs) have attracted considerable attention from both academia and industry due to their broad range of applications in edge devices. To obtain SLMs with strong performance, conventional approaches either pre-train the models from scratch, which incurs substantial computational costs, or compress/prune existing large language models (LLMs), which results in performance drops and falls short in comparison to pre-training. In this paper, we investigate the family of acceleration methods that involve both structured pruning and model training. We found 1) layer-wise adaptive pruning (Adapt-Pruner) is extremely effective in LLMs and yields significant improvements over existing pruning techniques, 2) adaptive pruning equipped with further training leads to models comparable to those pre-training from scratch, 3) incremental pruning brings non-trivial performance gain by interleaving pruning with training and only removing a small portion of neurons ($\sim$5\%) at a time. Experimental results on LLaMA-3.1-8B demonstrate that Adapt-Pruner outperforms conventional pruning methods, such as LLM-Pruner, FLAP, and SliceGPT, by an average of 1\%-7\% in accuracy on commonsense benchmarks. Additionally, Adapt-Pruner restores the performance of MobileLLM-125M to 600M on the MMLU benchmark with 200$\times$ fewer tokens via pruning from its larger counterparts, and discovers a new 1B model that surpasses LLaMA-3.2-1B in multiple benchmarks. The official code is released at \url{https://github.com/research4pan/AdaptPruner}.
\end{abstract}

\section{Introduction}
\label{sec:Introduction}

Large language models (LLMs)~\citep{KALYAN2024100048, openai2023gpt4} have demonstrated remarkable performance across a wide range of benchmarks. As their size increases, these models exhibit enhanced capabilities in understanding natural language and solving complex tasks through text generation~\citep{zhao2023survey}. However, achieving such performance requires models with billions of parameters, which presents significant challenges for practical deployment. The sheer scale of LLMs leads to high computational costs, making inference both resource-intensive and slow, and potentially introducing issues such as increased latency. Consequently, there is a growing demand for methods to compress LLMs~\citep{zhu2024surveymodelcompressionlarge}, aiming to reduce the number of parameters and improve inference speed, all while preserving the original model performance. Effective compression techniques hold the potential to create more efficient and deployable LLMs.

Several techniques have been proposed to compress LLMs, most of which fall into one of four categories: structured and unstructured pruning~\citep{cheng2024survey}, quantization~\citep{gholami2022survey}, low-rank factorization~\citep{sainath2013low}, and knowledge distillation~\citep{gou2021knowledge}. In this paper, we primarily focus on structured pruning, which can be combined together with further training to obtain strong Small Language Models. Structured pruning removes entire filters, layers, or specific model components from neural networks, enabling both compression and realistic acceleration. On top of that, it does not require specialized hardware or library support to achieve these benefits~\citep{he2023structured} in contrast to unstructured pruning.

\begin{figure*}[t!]
    \centering
    \includegraphics[width=\linewidth]{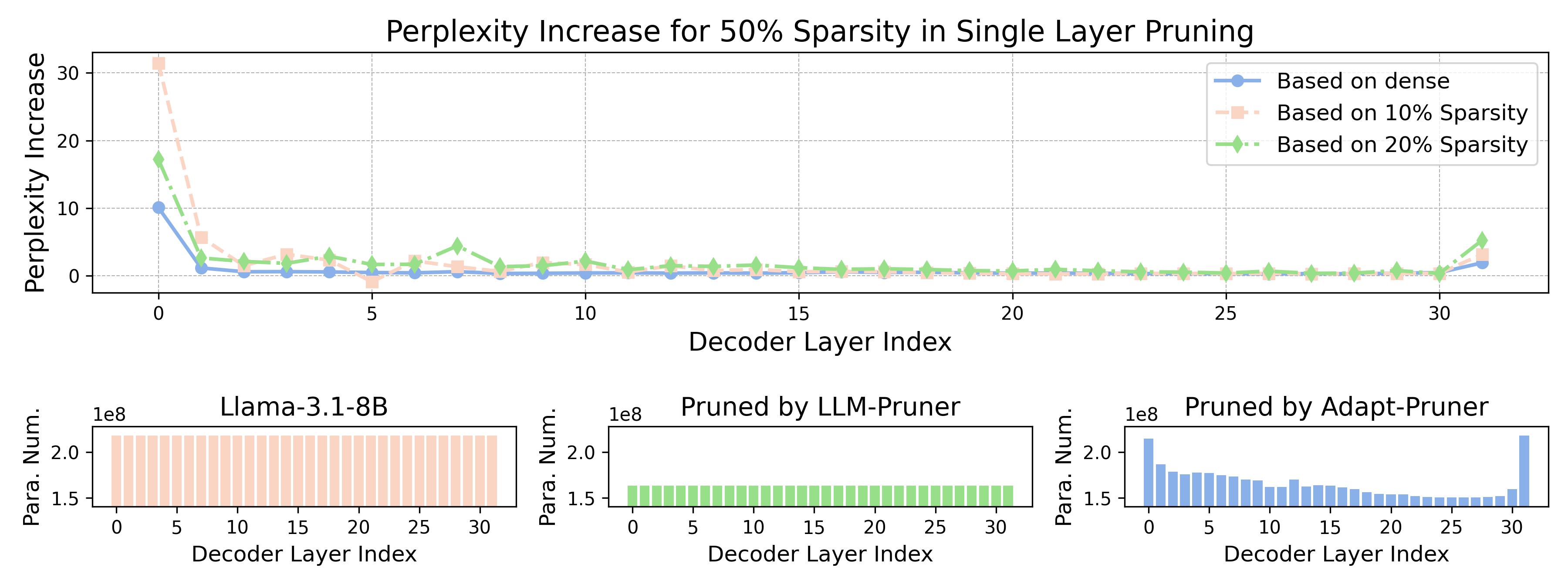}
    \caption{
        Layer sensitivity and pruned Models. The first row of figures shows the increase in perplexity when a single decoder layer is pruned at 50\% sparsity, compared to the dense LLaMA-3.1-8B model, as well as models uniformly pruned across all layers at 10\% and 20\% sparsity. The second row of figures illustrates the architecture of the pruned models, with each decoder layer represented by its corresponding number of parameters.
    }
    \label{fig:layer_sensitivity}
    \vspace{-5mm}
\end{figure*}

While many works on structured pruning focus on removing a fixed number of filters from weight matrices with minimal performance degradation, these methods often either skip important layers or apply uniform sparsity across all layers. However, as shown in Figure~\ref{fig:layer_sensitivity}, the importance of each decoder layer—and by extension, each weight matrix—varies significantly. To leverage this phenomenon, we introduce a novel approach called Adapt-Pruner. Unlike traditional pruning methods that enforce the same sparsity across all decoder layers, Adapt-Pruner operates in multiple steps. At each step, it calculates the relative importance of each decoder layer and applies varying sparsity levels, assigning higher sparsity to less important layers and lower sparsity to more critical ones. After determining the sparsity for each layer, the importance of each weight group is accessed using both magnitude and first-order information, leaving the least important groups for pruning.

Furthermore, when computational resources allow, structural pruning methods can be combined together with additional post-training stages, recovering the performance hurt by pruning. Multiple approaches~\citep{xia2024shearedllamaacceleratinglanguage,sreenivas2024llmpruningdistillationpractice} have investigated this fashion of training to obtain SLMs efficiently. Nonetheless, their methods fixate too much on the prune-the-train paradigm, which hinders them from further boosting the target SLM's performance. In contrast, a different paradigm is explored in this paper, offering new opportunities for structural-pruning-based acceleration methods.

Our main contributions are summarized as follows: 
\begin{enumerate} 
    \item A novel structured pruning method called Adapt-Pruner is proposed, which exploits the skewed importance distribution across LLM layers and significantly outperforms conventional pruning methods in commonsense benchmarks.

    \item A novel acceleration paradigm called Adapt-Accel is presented, which is the first method that interleaves the pruning with training in a highly frequent manner and demonstrates non-trivial improvements compared to past methods~\citep{xia2024shearedllamaacceleratinglanguage,sreenivas2024llmpruningdistillationpractice}.

    \item A novel family of models called Adapt-LLMs, which is obtained through Adapt-Accel, achieves superior performance over strong open-sourced models. In particular, Adapt-Pruner recovers the performance of MobileLLM~\citep{liu2024mobilellm} in MMLU~\citep{hendrycks2020mmlu} with $200\times$ less tokens via pruning from its larger counterparts. On top of that, a strong 1B model is discovered by pruning from DeepSeek-R1-Distill-Qwen-1.5B~\citep{deepseekr1}, outperforming Llama-3.2-1B~\citep{llama3modelcard} in multiple benchmarks, including MMLU, TruthfulQA~\citep{lin2021truthfulqa}, and AGIEval~\citep{zhong2023agieval}.
\end{enumerate}

\section{Related Work}
\label{sec:Related Work}

\paragraph{Pruning} Pruning removes weights and modifies the model's architecture. Formally, given a neural network $f(x;W)$, pruning produces a new model $f(x;M \odot W)$, where $M \in \{0,1\}^{|W|}$ is a binary mask that sets certain parameters to zero, and $\odot$ denotes element-wise multiplication. Pruning typically hurts the network's performance, so post-training is often employed to recover the loss~\citep{MLSYS2020_6c44dc73}. Overall, pruning methods can be categorized into two types, unstructured pruning and structured pruning. Unstructured pruning removes individual weights, resulting in sparsified weight matrices, but normally face difficulties in inference speedups when specialized hardware is unavailable~\citep{dery2024everybodyprunenowstructured}. In contrast, structured pruning operates at a larger granularity, removing entire weight groups. This includes width pruning~\citep{llm-pruner, SliceGPT}, which removes groups of coupled weights, and depth pruning~\citep{kim2024shortenedllamadepthpruning, siddiqui2024deeperlookdepthpruning}, which eliminates entire layers. Our focus is on post-training structured pruning, balancing generality and hardware efficiency.

\paragraph{Adaptive Sparsity for Pruning} Several works have explored adaptive compression.~\citet{dong2024prompt} selects transformer feedforward experts and removes feedforward neurons during inference based on their high activation magnitudes from input prompts. While this method is effective, we seek an approach that can reduce model size without depending on specific input prompts.~\citet{FLAP} computes the sample variance of each input feature and weights it by the squared norm of the corresponding column in the weight matrix to determine a layer’s importance and assign sparsity accordingly.~\citet{RLPruner} determines the optimal layer-wise sparsity distribution through reinforcement learning, where different sparsity patterns are sampled and updated based on their performance rewards. However, all aforementioned methods failed to take into account the loss on the functional aspect of models, where the overall mapping $X \rightarrow Y$ of the LLM is expected to be preserved during the pruning process to minimize the performance drop. Here $X$ and $Y$ are input and output tensors of the model. Our method utilizes the mapping information to assign an adaptive sparsity across different decoder layers.

\paragraph{Knowledge Distillation} Knowledge distillation is a technique used to transfer the advanced capabilities of high-performing LLMs to smaller models \citep{xu2024surveyknowledgedistillationlarge}. Combining knowledge distillation with pruning can yield strong performance, where the original model acts as the teacher and the compressed model serves as the student~\citep{sreenivas2024llmpruningdistillationpractice}.
Conventionally, knowledge distillation can be categorized into black-box~\citep{ho2022large,hsieh2023distilling} and white-box distillation~\citep{gu2023knowledge,latif2023knowledge,agarwal2023gkd,zhou2023distillspec,shum2024first}, depending on whether the model weights and prediction logits of the teacher model can be accessed. In that sense, structural-pruning-based acceleration methods can be roughly viewed as white-box knowledge distillation. However,
in this paper, we focus on the adaptive pruning algorithm with supervised fine-tuning only. Though it is expected to achieve even better performance when integrated with state-of-the-art knowledge distillation techniques in post-training, it normally incurs more complexity and training cost, hence we leave that for future exploration.

 \paragraph{Accelerating Small Language Model Training} Training small language models from scratch demands substantial computational resources, making pruning of larger models with recovery post-training an attractive alternative. Recent works have explored various approaches to this challenge.~\citet{xia2024shearedllamaacceleratinglanguage} proposes a systematic pruning framework that optimizes across four architectural dimensions (layer count, hidden dimension, attention heads, and MLP size) towards target architectures, followed by a continual post-training phase phase with dynamic batch loading.~\citet{sreenivas2024llmpruningdistillationpractice} leverages knowledge distillation, first fine-tuning the original model as a teacher and then transferring its knowledge to the pruned student model. However, these methods rely on single-phase post-training and do not exploit the potential benefits of incremental pruning with interleaved recovery phases, which is demonstrated to be effective in the experiments.

\section{Method}
\label{sec:Method}

\subsection{Adapt-Pruner: Layer-wise Adaptive Pruning aim for Mapping-Preserving}
Given a large language model $\mathcal{M}$, represented as a sequence of embedded layers with $\mathcal{N}$ decoder layers, denoted as $\mathcal{L}^\mathcal{N}$, along with a final output layer, our method leverages the insight that each decoder layer contributes differently to the model's overall performance. Furthermore, the contribution of each layer is measured by its importance of maintaining the original functional mapping of the model after pruning.

\begin{figure*}[t!]
    \centering
    \includegraphics[width=\linewidth]{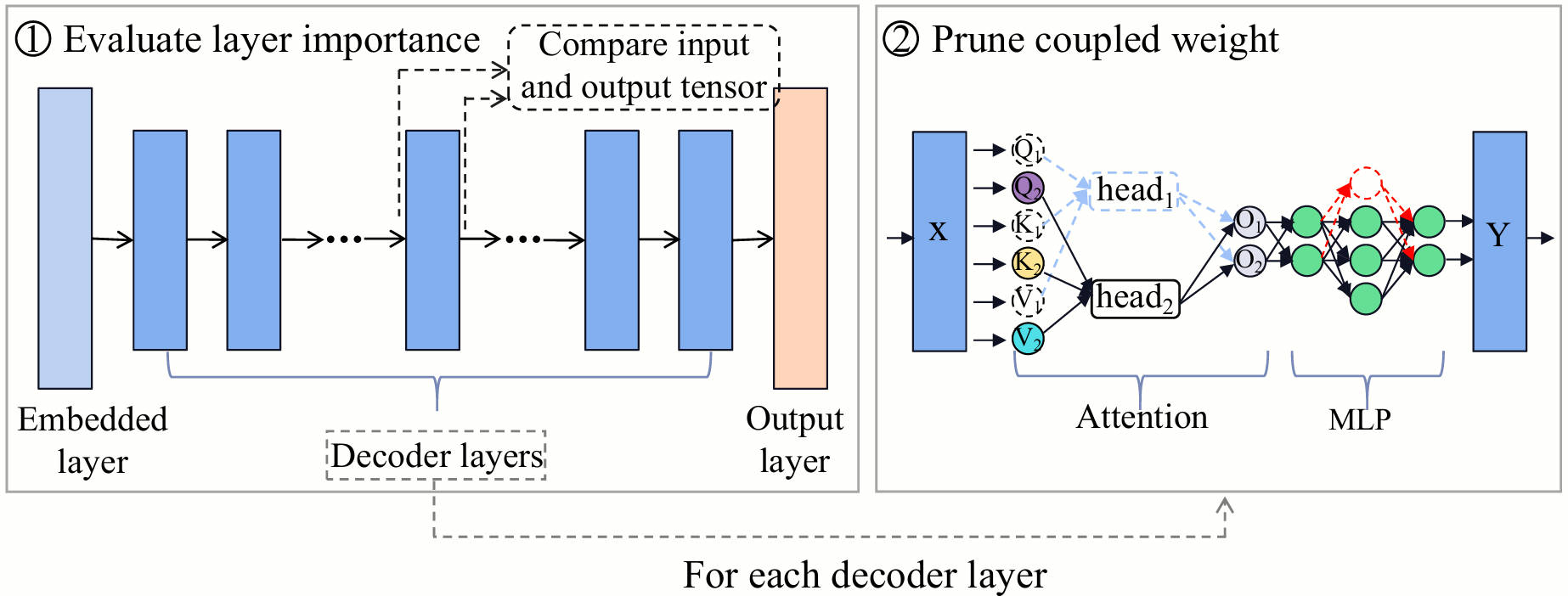}
    \caption{
        Adapt-Pruner: measuring the distance between each decoder layer's input and output tensors to assess its importance and assigning a corresponding sparsity. Based on this assigned sparsity, the coupled weights in each decoder layer are pruned accordingly.
    }
    \label{fig:adapt_pruner}
    \vspace{-5mm}
\end{figure*}

Specifically, Adapt-Pruner compresses the model through multiple iterations, with each iteration comprising two steps:
\begin{enumerate}
    \item \textbf{Evaluating layer importance}: Quantitatively computing the importance of each decoder layer and assigning a corresponding pruning sparsity.
    \item \textbf{Pruning coupled weights}: Grouping the weights within each decoder layer, evaluating the importance of each coupled structure, and pruning the least important structures based on the assigned sparsity
\end{enumerate}
Figure \ref{fig:adapt_pruner} gives an illustration of our method.

\subsubsection{Assign Sparsity based on Importance}

Let $\mathcal{L}^i$ denote the $i$-th decoder layer, $I^i$ and $\mathcal{S}^i$ represent the importance and sparsity of the $i$-th decoder layer, and $\mathcal{L}^{i}_{in}$ and $\mathcal{L}^{i}_{out}$ denote the input and output tensors of the $i$-th decoder layer, respectively. Our goal is to estimate the importance of each decoder layer.

\paragraph{Estimate Decoder Layer's Importance} Our pruning method targets only the multi-head attention and multilayer perceptron components within the self-attention layers, leaving the hidden size unchanged. Consequently, the input and output tensors for each decoder layer have identical shapes:
\begin{align} 
\forall i = 0,1,\dots,N-1, \text{Shape}(\mathcal{L}^{i}_{in}) &= \text{Shape}(\mathcal{L}^{i}_{out}) \\
&= (\mathcal{B}, \mathcal{L}, \mathcal{H}) 
\end{align}
where $\mathcal{B}, \mathcal{L}, \mathcal{H}$ denote the batch size, sequence length, and hidden size, respectively. Based on this, we use a function that measures the vector similarity or distance between $\mathcal{L}^{i}_{in}$ and $\mathcal{L}^{i}_{out}$ to assess the changes in the tensor caused by each decoder layer. The greater the similarity or the smaller the distance between $\mathcal{L}^{i}_{in}$ and $\mathcal{L}^{i}_{out}$, the less important that decoder layer is. This intuition derives from~\citet{li2024mix}, where it is observed that top self-attention layers have diminished gradient norms and those layers serve similar purposes as identity functions.

A practical choice for this distance measurement is cosine similarity, where a decoder layer's importance is computed as follows:
\begin{align} 
\forall i = 0,1,\dots,N-1, I^i = -\text{cosine\_similarity}(\mathcal{L}^{i}_{in}, \mathcal{L}^{i}_{out})
\end{align}
This method can easily be extended to alternative similarity or distance functions, such as Euclidean or Manhattan distance. To ensure consistency, we normalize the decoder layer's importance to the range $[-1,1]$ with a mean of 0, as follows:
\begin{align} 
I^i \leftarrow I^i - I_{\text{mean}} \\
I^i \leftarrow \frac{I^i}{\max |\text{abs}(I)|} 
\end{align}

\paragraph{Assign Sparsity} After obtaining the importance of each layer, an ad-hoc approach is adopted to link a layer's importance to its sparsity, which decides the number of neurons it will be pruned. Let $A$ being any constants, the targeted sparsity $\mathcal{S}^i$ for each layer $i$ can be:
\begin{align} 
\label{link_sparsity_importance}
\forall i = 0,1,\dots,N-1, \mathcal{S}^i = \mathcal{S}_{base} - A \cdot I^i 
\end{align}
where $\mathcal{S}_{base}$ is the targeted overall sparsity of the model. This formula ensures that each decoder layer's sparsity is inversely proportional to its importance, and the averaged sparsity is consistent with the intended overall model sparsity. We call the hyperparameter $A$ as the amplitude of sparsity.

In addition, since the importance distribution of each layer varies throughout the pruning process according to the observations from Figure~\ref{fig:layer_sensitivity}, progressive adjustment during pruning becomes necessary for good performance. This leads to the multi-stage-pruning design in Adapt-Pruner.

\begin{algorithm}[t]
\small
  \caption{Adaptive Pruning Algorithm}\label{alg:adapt_pruning}
  \begin{algorithmic}[1]
    \REQUIRE Number of decoder layer in the LLM $N$, decoder layer instances in the LLM $\{\mathcal{L}^i\}_{i=1}^N$, overall sparsity after pruning $\mathcal{S}$, iteration times to prune $T$, pruning ratio threshold to apply post-train $P$, training data $\mathcal{D}$, amplitude of sparsity between decoder layers $A$, similarity function, use cos in default $Sim_{func} \gets \text{cosine\_similarity}()$
    \vspace{2mm}
    \FOR{$i \gets 1\ldots T$}
      \STATE $S_{cur} \gets (S \cdot i / T)$
      \STATE $I \gets \{0\}^N$, $S \gets \{0\}^N$
      \FOR{$j \gets 1\ldots N$}
        \STATE $I^j \gets Sim_{func}(\mathcal{L}^j_{in}, \mathcal{L}^j_{out})$
      \ENDFOR
      \STATE Normalize $I^N$ to have 0 mean value, limit range to [-1, 1] and times -1 if lower is better
      \FOR{$j \gets 1\ldots N$}
        \STATE $S^j \gets S_{cur} - A\cdot I^j$
      \ENDFOR
      \WHILE{Current sparsity $ > P$}
        \STATE Adaptively prune the LLM based on $S_j$
      \ENDWHILE
    \ENDFOR
  \end{algorithmic}
\end{algorithm}


\subsubsection{Pruning Weight Groups Inside Decoder Layer}

To enable structure-aware pruning in Adapt-Pruner, methods from \citet{llm-pruner, depgraph} are employed to build dependency graphs for LLMs, which facilitates automatic identification and extraction of coupled structures in LLMs. 

\paragraph{Weight Group Importance} With coupled structures and target sparsity defined for each group, the next step is selecting weight matrices to prune with minimized performance degradation. For any grouped weight structure $\mathcal{G} = {W^k}$, containing $k$ weight matrices, a calibration dataset $\mathcal{D}$ is adopted to assess the relative importance of each matrix. Following \citep{lecun1989optimal, llm-pruner}, the importance of the $i$-th weight matrix in layer $\mathcal{L}$ is defined as:
\begin{align}
I_{W_i} &= | \Delta \mathcal{L}(\mathcal{D})| \\ 
&= |\mathcal{L}_{W_i}(\mathcal{D}) - \mathcal{L}_{W_i=0}(\mathcal{D})| \\
&=\left| {\frac{\partial \mathcal{L}^{\top}(\mathcal{D})}{\partial W_i} W_i}-\frac{1}{2} {W_i}^{\top} H W_i + \mathcal{O}\left(\| W_i \|^3\right) \right|
\end{align}
where $H = \frac{\partial \mathcal{L}^2(\mathcal{D})}{\partial W_i^2}$ is the Hessian matrix. Calculating the Hessian requires $\mathcal{O} (N^2)$ computational resources, so only first-order terms are retained for acceleration purposes. This simplifies the estimated weight matrix importance to:
\begin{align}
\hat{I}_{W_i} =\left| {\frac{\partial \mathcal{L}^{\top}(\mathcal{D})}{\partial W_i} W_i}\right|
\end{align}
Thus, each weight matrix's importance can be approximated by taking the $l_1$ norm of the element-wise product between its gradient (derived from the calibration dataset) and its weight value. After computing importance scores, the matrices are sorted, where the ones with the lowest scores are pruned to achieve the desired sparsity level. The complete pruning procedure is detailed in Algorithm \ref{alg:adapt_pruning}.

\subsection{Adapt-Accel: Incremental Pruning with Interleaved Recovery Training} To achieve the target overall sparsity, our model undergoes multiple rounds of pruning, which inevitably leads to performance degradation. Inspired by neuroplasticity in biological neural networks~\citep{walloe2014stereological}, periodic interleaved recovery phases are introduced through post-training after each pruning. The optimal pruning ratios for triggering these recovery phases were determined through ablation studies in Section~\ref{sec:exp:ablation}.

\begin{figure*}[t!]
    \centering
\includegraphics[width=0.95\textwidth]{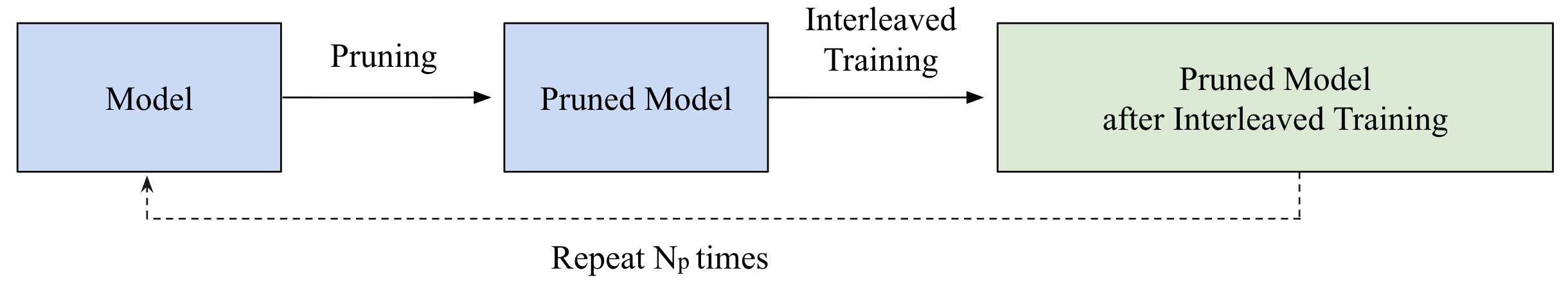}
    \\
    \caption{Adapt-Accel: Incremental pruning with interleaved training. $N_P$ number of interleaves are adopted in the whole process. Given a model with size $|\mathcal{L}_{\text{large}}|$ and target size $|\mathcal{L}_{\text{small}}|$, this leads to an incremental pruning ratio of $P = (|\mathcal{L}_{\text{small}}| / |\mathcal{L}_{\text{large}}|)^{1/N_P}$ each time, where the training set is randomly split into $N_P$ subsets for $N_P$ interleaved trainings separately. Notice that the number of training samples gradually increases according to Algorithm~\ref{alg:adapt_pruning}, as more important weights are expected to be pruned in later phases.}
    \label{fig:adapt-accel-illust}
\end{figure*}

In addition, it is observed that different pruning phases lead to different levels of performance deterioration. Specifically, pruning in later phases is more likely to remove important neurons. Based on this intuition, a linear growth schedule is introduced for training data allocation: For $i$-th post-training of $T$ total pruning and post-training iterations, $|D_i|$ tokens are sampled from $|D|$ total training tokens where 
\begin{align}
|D_{i}| = \frac{2 (i+1)}{|D|(|D|+1)}
\end{align}
This linear growth schedule for data allocation serves two key purposes:
\begin{itemize}
    \item It enables more frequent parameter updates during later phases when the model requires more extensive recovery. \item And it preserves knowledge acquired in earlier phases from being pruned in subsequent iterations.
\end{itemize}
This recovery phase is further optimized by splitting and distributing the learning rate schedule across post-training phases, ensuring efficient restoration of model performance after each pruning step.

\section{Experiment}
\label{sec:Experiment}

\subsection{Adapt-Pruner as Effective LLM Pruners}
\label{sec:exp:pruner}

Adapt-Pruner exploits the skewness of importance across layers in a mapping-preserved manner, which allows the pruning process to automatically identify prunable layers that least affect the functionality of the target LLM.

\begin{table*}[!b]
    \centering
    \caption{\textbf{Pruning-only}: Structured pruning methods across different sparsity levels in Llama-3.1-8B over three trials.} 
    \vspace{0.05in}
    \label{table:main:pruner}
    \resizebox{\linewidth}{!}{
    \begin{tabular}{ll|ccccccc|cr}
        \toprule
        \toprule
        Model & Method & ARC-e & ARC-c & HellaSwag & OBQA & PIQA & SIQA & Winogrande & Average$\uparrow$ & WikiText2 ppl. $\downarrow$ \\
        \midrule
        
        \multirow{4}{*}{\parbox{1.8cm}{Ratio = 20\% \  w/o tune}} & LLM-Pruner & 63.92 & 39.53 & 64.85 & 38.47 & 76.64 & 42.80 & 62.85 & 55.58 & 14.78 \\
        & FLAP & 61.01 & 37.37 & 58.43 & 35.40 & 73.27 & 43.40 & 64.46 & 53.33 & 16.73 \\
        & SliceGPT & 52.27 & 29.01 & 56.37 & 32.00 & 69.84 & 41.85 & 59.46 & 48.69 & 19.61 \\
        & Adapt-Pruner & 66.43 & 39.65 & 66.71 & 38.93 & 76.87 & 44.10 & 66.19 & \bf 56.98 & \bf 14.54 \\
        \cmidrule{1-11}
        \multirow{4}{*}{\parbox{1.8cm}{Ratio = 40\% \  w/o tune}} & LLM-Pruner & 33.90 & 21.79 & 30.00 & 25.40 & 57.02 & 35.45 & 51.30 & 36.41 & 162.81 \\
        & FLAP & 25.53 & 26.25 & 26.49 & 25.73 & 52.36 & 34.12 & 50.59 & 34.44 & 6987.58 \\
        & SliceGPT & 33.94 & 21.79 & 33.72 & 26.27 & 57.73 & 36.44 & 49.72 & 37.09 & 85.60 \\
        & Adapt-Pruner & 45.16 & 25.97 & 44.88 & 30.40 & 66.74 & 39.03 & 56.75 & \bf 44.13 & \bf 33.75 \\
        \cmidrule{1-11}
        \multirow{4}{*}{\parbox{1.8cm}{Ratio = 60\% \  w/o tune}} & LLM-Pruner & 26.94 & 24.43 & 26.67 & 27.93 & 51.22 & 34.14 & 49.43 & 34.39 & 2501.76 \\
        & FLAP & 26.60 & 26.53 & 26.00 & 27.07 & 51.63 & 33.95 & 49.62 & 34.49 & 141572.73 \\
        & SliceGPT & 28.95 & 21.61 & 28.12 & 26.13 & 53.01 & 34.49 & 49.01 & 34.48 & 218.96 \\
        & Adapt-Pruner & 32.49 & 23.64 & 30.84 & 26.40 & 56.66 & 35.59 & 49.43 & \bf 36.44 & \bf 119.95 \\
        \bottomrule
        \bottomrule
    \end{tabular}
    }
\end{table*}

\paragraph{Setup} To demonstrate the effectiveness of Adapt-Pruner, different pruners are evaluated and compared on Llama-3.1-8B \citep{llama3modelcard}. For task-agnostic performance evaluation of the pruned models, zero-shot classification is performaned on popular common-sense reasoning datasets: ARC-easy and ARC-challenge \citep{arc}, HellaSwag \citep{hellaswag}, OpenBookQA \citep{OpenBookQA}, PIQA \citep{piqa}, SIQA \citep{siqa} and WinoGrande \citep{winogrande}, where the averaged accuracy is reported. In particular, length-normalized accuracy is reported for any benchmarks that require length-dependent accuracy. Additionally, we supplement our evaluation with a generation task using WikiText2 \citep{wikitext2}. Following prior work \citep{llm-pruner, FLAP, SliceGPT}, we employ the LM Evaluation Harness \citep{eval-harness} with default parameters, except that all models use the bfloat16 data type, and the batch size is set to `auto' during evaluation.

\paragraph{Results}

As shown in Table~\ref{table:main:pruner}, Adapt-Pruner outperforms all baselines by a non-trivial margin. In particular, Adapt-Pruner excels at preserving commonsense knowledge in LLMs during the pruning process. This is highly relevant to the skewed knowledge distribution across LLM layers, where the bottom layers are observed to be more important than the top layers~\citep{pan2024lisa,li2024mix}.

\begin{wraptable}{r}{0.45\textwidth}
\centering
\vspace{-0.28in}
\caption{\textbf{Dataset} for training or interleaved training after pruning.}
\vspace{0.05in}
\begin{tabular}{@{}lr@{}}
\toprule
\toprule
\textbf{Dataset}                    & \textbf{\#Tokens} \\ \midrule
\texttt{Open-Orca/OpenOrca}                   & 1.80B            \\
\texttt{allenai/WildChat-1M}                  & 0.66B                \\
\texttt{lmsys/lmsys-chat-1m}                  & 0.49B               \\
\texttt{teknium/OpenHermes-2.5}               & 0.49B                \\
\texttt{HuggingFaceH4/ultrachat\_200k}        & 0.20B               \\
\texttt{openbmb/UltraInteract\_sft}           & 0.18B               \\

\texttt{O1-OPEN/OpenO1-SFT}                   & 74M              \\ 
\texttt{yahma/alpaca-cleaned}                 & 12M              \\
\texttt{databricks/databricks-dolly-15k}      & 3.6M                 \\
\midrule
\textbf{Total}                     & \textbf{3.87B}        \\ \bottomrule \bottomrule
\end{tabular}
\vspace{-0.2in}
\label{tab:main:dataset_info}
\end{wraptable}

\subsection{Adapt-Accel as Efficient LLM Trainers}
\label{sec:exp:acceleration}

Built on top of Adapt-Pruner, Adapt-Accel incorporates interleaved pruning and training to accelerate the optimization of SLMs, which is shown to be a better strategy compared to past methods~\citep{xia2024shearedllamaacceleratinglanguage,sreenivas2024llmpruningdistillationpractice}.

\paragraph{Setup}

As the training of language models requires a non-trivial amount of computational resources, a smaller family of models are adopted for this section of experiments. MobileLLM~\citep{liu2024mobilellm} is a series of SLMs developed by Meta for on-device deployment purposes and stands for one of the strongest SLMs on the scale of 125M/350M.

\begin{wraptable}{r}{0.45\textwidth}
    \centering
    \vspace{-0.28in}
    \caption{\textbf{Pruning + Training}: Comparison of different structural-pruning-based acceleration methods on MobileLLM-350M $\rightarrow$ 125M.} \label{table:main:acceleration}
    \vspace{0.05in}
    \resizebox{\linewidth}{!}{
    \begin{tabular}{l|ccc}
        \toprule
        \toprule
        Method & \makecell{BBH\\(3-shot)} & \makecell{TruthfulQA\\(1-shot)} & \makecell{AGIEval\\(0-shot)} \\
        \midrule
        \makecell[l]{ShearedLLaMA\\~\cite{xia2024shearedllamaacceleratinglanguage}} & 12.29 & 34.88 & 30.53 \\
        \midrule
        \makecell[l]{Minitron\\~\citep{sreenivas2024llmpruningdistillationpractice}} & 12.56 & 34.13 & 30.36 \\
        \midrule
        Adapt-Accel & \bf 13.75 & \bf 36.18 & \bf 31.34 \\
        \bottomrule
        \bottomrule
    \end{tabular}
    }
    \vspace{-0.2in}
\end{wraptable}

To demonstrate the superiority of Adapt-Accel, three benchmarks are employed for evaluation, including BBH~\citep{srivastava2022bbh}, TruthfulQA~\citep{lin2021truthfulqa}, AGIEval~\citep{zhong2023agieval}, which assess all methods performance in different aspects beyond commonsense reasoning. A hybrid dataset with 3.87B tokens is utilized for training, where the data source is available in Table~\ref{tab:main:dataset_info}.

For the interleaved training in Adapt-Accel, a total number of $N_P = 20$ interleavings are adopted in Adapt-Accel, which leads to a pruning ratio of $(125 / 300)^{1/N_P} \approx 95.7\%$ per training. In other words, the pruning and training will be applied alternatively for $20$ times, where each pruning will remove $\sim 95.7\%$ of the current models' weights, along with a follow-up recovery training in a $|D_i| = 2(i+1)/(|D|^2 + |D|)$ random samples (without replacement) from the training set at the $i$-th iteration.

\paragraph{Results}

As shown in Table~\ref{table:main:acceleration}, Adapt-Accel outperforms both ShearedLLaMA~\citep{xia2024shearedllamaacceleratinglanguage} and NVIDIA's Minitron Approach~\citep{sreenivas2024llmpruningdistillationpractice} by a non-trivial margin. This demonstrates that the adaptively pruned models have preserved essential components in the original SLMs, and are still capable of learning new knowledge from the training set effectively.

\subsection{Adapt-LLMs as Strong LLMs}

Adapt-Accel is a favorable tool for fast and flexible customization of model sizes depending on the practical use cases. Specifically, once the large version of LLMs has been obtained from pre-training or other sources, Adapt-Accel can be utilized to inherit the capabilities from the target LLM and accelerate the training of its smaller versions. The family of SLMs obtained in this fashion, named Adapt-LLMs, not only reduces costs during its training process, but also exhibits significant performance improvements.

\paragraph{Setup}
To provide evidence in support of the claimed strengths of Adapt-LLM, two types of experiments are conducted, individually demonstrating the acceleration and performance benefits of Adapt-LLM. The first branch of experiments focuses on the acceleration aspect of Adapt-Accel, where different sizes of MobileLLMs, ranging from 350M to 1B, are employed. The second branch of experiments emphasizes performance, where Adapt-Accel is applied to Qwen-2.5-0.5B~\citep{qwen2.5} and Deepseek-R1-Distill-Qwen-1.5B~\citep{deepseekr1}, proving that the pruned Adapt-LLMs can still match or even surpass popular strong open-source models without the heavy cost of pretraining. The same benchmarks and datasets of Section~\ref{sec:exp:acceleration} are adopted here.

\begin{table*}[t!]
    \centering
    \caption{\textbf{Pruning + Training}: Recovering MobileLLM's performance with much $200\times$ less tokens. Here MobileLLM-X $\rightarrow$ Y means to prune the model from size X to size Y with Adapt-Accel.} \label{table:main:model_mobilellm_recover}
    \vspace{0.05in}
    \resizebox{\linewidth}{!}{
    \begin{tabular}{l|r|cccc}
        \toprule
        \toprule
        Model & $\#$Training Tokens & \makecell{BBH (3-shot)} & \makecell{TruthfulQA (1-shot)} & \makecell{AGIEval (0-shot)} &
        \makecell{MMLU (5-shot)} \\
        \midrule
        \makecell[l]{MobileLLM-125M} & $\sim 10^{12}$  & \bf 18.45 & 32.88 & 31.19 & 24.65
        \\
        \makecell[l]{MobileLLM-350M $\rightarrow$ 125M} & $3.87 \times 10^9$ & 13.75 & \bf 36.18 & \bf 31.34 & \bf 25.20       
        \\
        \midrule
        \makecell[l]{MobileLLM-350M} & $\sim 10^{12}$  & 19.98 & 29.86 & 30.85 & 26.11
        \\
        \makecell[l]{MobileLLM-600M $\rightarrow$ 350M} & $3.87 \times 10^9$ & \bf 23.61 & \bf 34.22 & \bf 31.93 & \bf 32.50       
        \\
        \midrule
        \makecell[l]{MobileLLM-600M} & $\sim 10^{12}$  & 23.99 & 29.29 & 30.53 & 26.46
        \\
        \makecell[l]{MobileLLM-1B $\rightarrow$ 600M} & $3.87 \times 10^9$ & \bf 26.60 & \bf 32.35 & \bf 34.24 & \bf 36.77      
        \\
        \bottomrule
        \bottomrule
    \end{tabular}
}
\end{table*}

\paragraph{Results}

As shown in Table~\ref{table:main:model_mobilellm_recover}, Adapt-LLM pruned from the larger version of MobileLLMs recovers its performance across all model sizes in all benchmarks for models larger than 350M, at a reduced cost of $200\times$ less training tokens. This offers strong evidence that Adapt-Accel is a promising acceleration technique especially suitable for customizing model sizes flexibly.

Table~\ref{table:main:model_better} further demonstrates the performance gain brought by Adapt-Accel via pruning from strong LLMs. It is worth noticing that Deepseek-R1-Distill-Qwen-1.5B~\citep{deepseekr1} $\rightarrow$ 1B leads to a 1B model even stronger than LLaMA-3.2-1B~\citep{llama3modelcard} in MMLU, delivering three orders of magnitude cost reduction in terms of training tokens. This indicates that Adapt-Accel can serve as a favorable tool for inheriting performance from strong open-source LLMs, allowing researchers and engineers with limited computational resources to still keep up with the fast iteration speed of state-of-the-art LLMs.

\begin{table*}[b!]
    \centering
    \caption{\textbf{Pruning + Training for model-size customization}: Surpassing popular open-sourced SLMs.} \label{table:main:model_better}
    \vspace{0.05in}
    \resizebox{\linewidth}{!}{
    \begin{tabular}{l|r|cccc}
        \toprule
        \toprule
        Model & $\#$Training Tokens & \makecell{BBH (3-shot)} & \makecell{TruthfulQA (1-shot)} & \makecell{AGIEval (0-shot)} &
        \makecell{MMLU (5-shot)} \\
        \midrule
        \makecell[l]{MobileLLM-350M} & $\sim 10^{12}$ & 19.98 & 29.86 & 30.85 & 26.11   
        \\
        \makecell[l]{Qwen-2.5-0.5B $\rightarrow$ 350M} & $3.87 \times 10^9$ & \bf 26.97 & \bf 34.86 & \bf 33.23 & \bf 38.48    
        \\
        \midrule
        \makecell[l]{MobileLLM-1B} & $\sim 10^{12}$  & 25.88 & 30.26 & 31.34 & 24.98
        \\
        \makecell[l]{LLaMA-3.2-1B} & $\sim 9 \times 10^{12}$  & \bf 31.01 & 30.41 & 31.68 & 31.27
        \\
        \makecell[l]{Deepseek-R1-Distill-Qwen-1.5B $\rightarrow$ 1B} & $3.87 \times 10^9$  & \underline{30.46} & \bf 37.50 & \bf 34.34 & \bf 34.62
        \\
        \bottomrule
        \bottomrule
    \end{tabular}
    }
\end{table*} 

\paragraph{Computation Cost} An 8B-sized model can be pruned adaptively in 2 to 5 minutes on a single NVIDIA A40 GPU, and in 15 to 30 minutes on an Intel(R) Xeon(R) Gold 6346 CPU. After applying iterative post-training, the full compression process takes between 3 and 18 hours. Benchmark evaluation of the compressed models requires an additional 15 to 30 minutes. In the experiments of Qwen2.5-0.5B $\rightarrow$ 350M, the training process costs $\sim$ 72 GH200 GPU hours. Compared with the $\sim$ 4608 A100 GPU hours pretraining cost of MobileLLM-350M, it is at least 15$\times$ more efficient.

\subsection{Ablation Study: Optimal Interleaving Frequency}
\label{sec:exp:ablation}

\begin{figure*}[h!]
    \centering
    \includegraphics[width=0.45\textwidth]{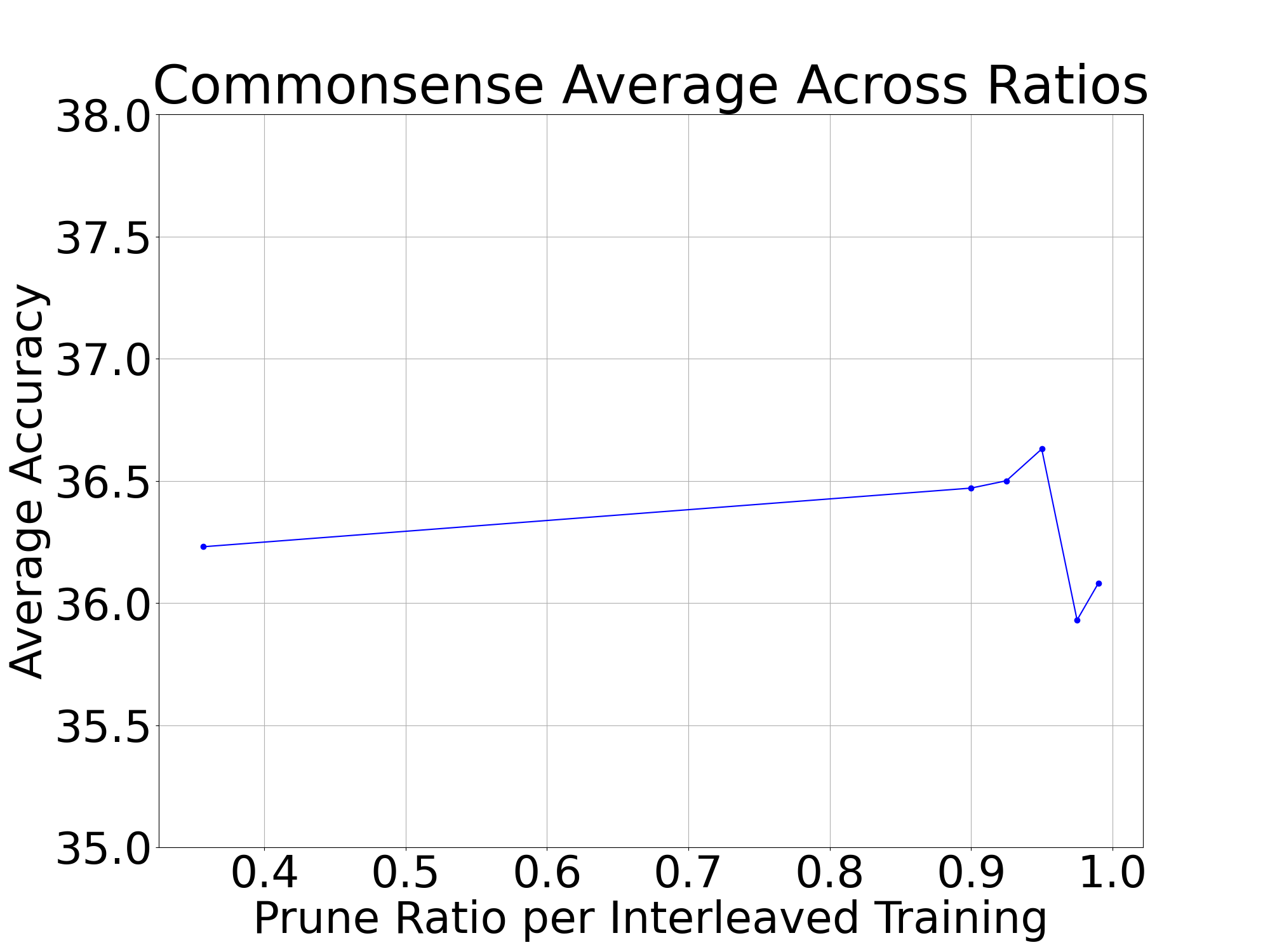}
    \includegraphics[width=0.45\textwidth]{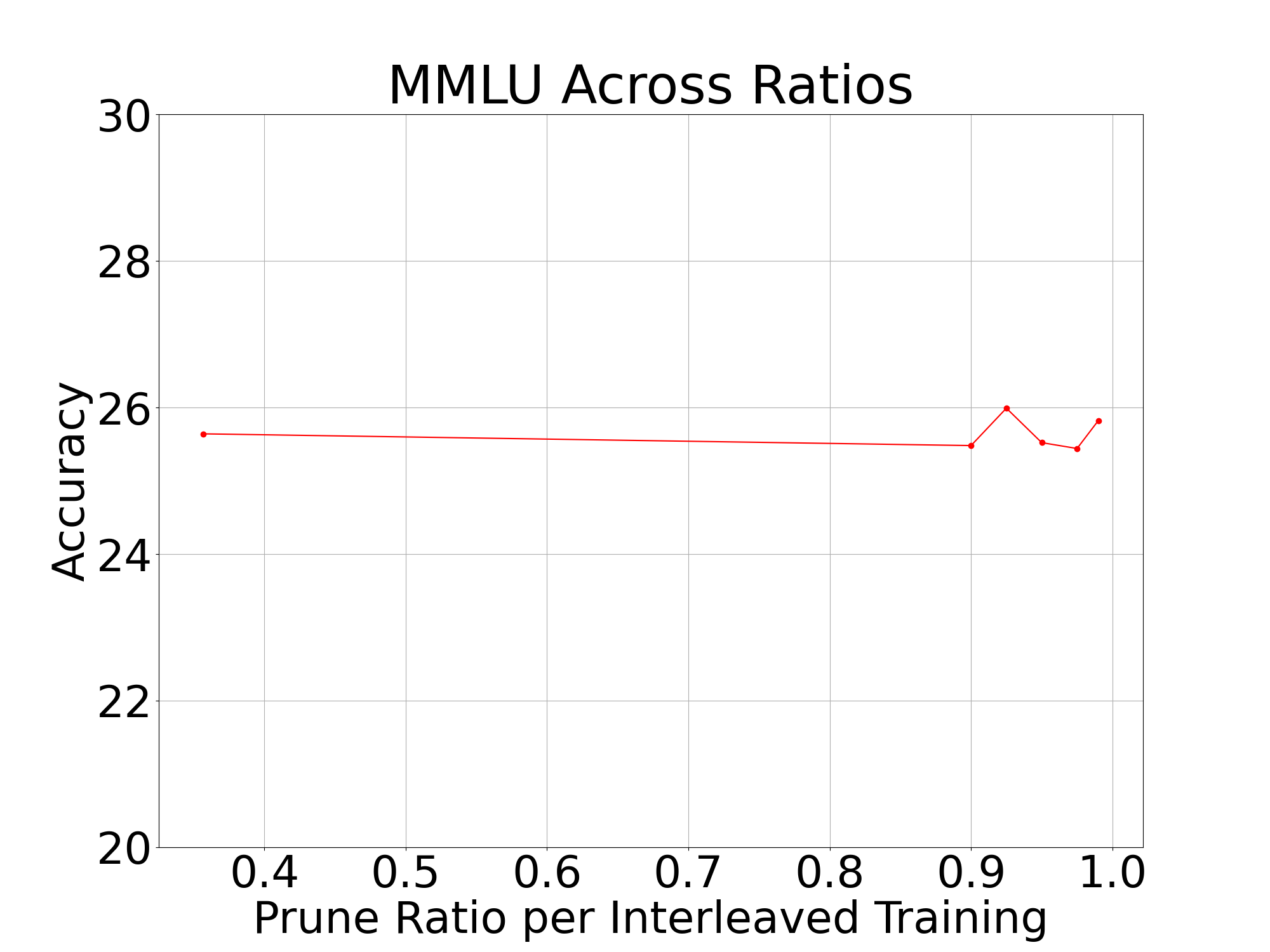}
    \caption{\textbf{Ablation studies} over pruning ratio per interleaved training, which shows the optimal value is $\sim 95\%$, meaning it is best to interleave the training of SLM after every $\sim 5\%$ weight/neuron removals.}
    \label{fig:ablation_interleaved_training}
\end{figure*}

Interleaved training is shown to be quite beneficial compared to the traditional prune-then-train paradigm. To further investigate the optimal frequency for interleaved training during the pruning process, additional experiments are conducted on MobileLLM-350M$\rightarrow$125M.

\paragraph{Setup}
All experiments are conducted on MobileLLM-350M, with 1B tokens sampled from Slimpajama~\citep{slimpajama} and 1/6 random samples in the aforementioned dataset (Table~\ref{tab:main:dataset_info}). Two types of benchmarks are adopted, including the commonsense benchmarks in Section~\ref{sec:exp:pruner} and MMLU benchmark in Section~\ref{sec:exp:acceleration}. Pruning ratios per training, ranging from $\{0.36, 0.90, 0.925, 0.95, 0.975, 0.99\}$, are searched to decide the optimal value, which corresponds to the number of interleaves $N_P \in \{1, 9, 12, 20, 38, 96\}$ separately.

\paragraph{Results}

As shown in Figure~\ref{fig:ablation_interleaved_training}, the optimal interleave frequency is around $95\%$ pruning ratio per training, i.e. every recovery training after $\sim 5\%$ removal of weights or neurons. It is worth noticing that the performance degrades significantly when the interleave frequency is too low or too high, implying the occurrence of knowledge loss caused by large-portion pruning, or unstable learning due to too-frequent pruning. This phenomenon is quite intriguing as it resembles the disappearance of neurons in human brains.

\section{Conclusion}
\label{sec:Conclusion}

In this paper, a novel family of structural pruning methods called AdaptPrune is proposed for LLMs. Adapt-Pruner is motivated by 1) the skewness of importance across decoder layers and 2) the goal of preserving the input-output mapping for all the layers. These two properties of AdaptPruner lead to significant improvement over conventional pruning methods.

On top of that, Adapt-Pruner gives rise to a novel acceleration paradigm called Adapt-Accel, which is the first acceleration approach that combines Adapt-Pruner with interleaved training, a technique shown to provide non-trivial performance gain over the traditional prune-then-train framework. Adapt-Accel provides consistent improvement over past structural-pruning-based acceleration methods, including ShearedLLaMA and Minitron.

Adapt-Accel further enables efficient and flexible customization of model sizes by pruning from larger-sized LLMs. Specifically, it is capable of recovering MobileLLMs' performance from their larger counterparts, and discovers a 1B-sized model with better performance than LLaMA-3.2-1B in multiple benchmarks.

\section*{Impact Statement}

This paper presents work whose goal is to advance the field
of Machine Learning. Specifically, we expect the proposed methods can greatly reduce the pre-training cost of SLMs, leading to less computational resource consumption, less carbon dioxide emissions, and faster development of SLMs.

\bibliographystyle{unsrtnat}
\bibliography{references}  






\newpage
\appendix
\onecolumn

\section{Experimental Details}
\label{appendix:exp_details}


\paragraph{Basic Setup} We extend the LLM-Pruner framework \citep{llm-pruner, depgraph} as our baseline, which incorporates modules for computing similarity scores and adaptive pruning using PyTorch \citep{paszke2019pytorch}. Our experiments utilize both NVIDIA GH200 and H100 GPUs, where a single GPU (either GH200 or H100) is utilized for pruning and evaluation tasks while 4 additional GPUs are employed in parallel for post-training. The metrics and the number of shots used for each benchmark are available in Table~\ref{tab:appendix:metrics_details}.

\begin{table}[h!]
\centering
\caption{Details for Evaluation Benchmarks.}
\label{tab:appendix:metrics_details}
\vspace{0.05in}
\begin{tabular}{@{}lrr@{}}
\toprule
\toprule
\textbf{Benchmark}                    & \textbf{Metric} & \textbf{n-shot} \\ \midrule
\texttt{ARC-e}                   & acc\_norm & 0            \\
\texttt{ARC-c}                  & acc\_norm & 0                \\
\texttt{BoolQ}                  & acc  & 0             \\
\texttt{HellaSwag}                  & acc\_norm  & 0             \\
\texttt{OBQA}                  & acc\_norm  & 0             \\
\texttt{PIQA}                  & acc\_norm  & 0             \\
\texttt{SIQA}                  & acc  & 0             \\
\texttt{Winogrande}                  & acc  & 0             \\
\texttt{WikiText2}                  & word\_perplexity  & 0             \\
\texttt{BBH}                  & exact\_match  & 3             \\
\texttt{TruthfulQA}                  & acc  & 1             \\
\texttt{AGIEval}                  & acc\_norm  & 0             \\
\texttt{MMLU}                  & acc  & 5             \\
\bottomrule \bottomrule
\end{tabular}
\label{tab:main:bencmark_info}
\end{table}

\paragraph{Experimental Settings for Adapt-Pruner} For Adapt-Pruner, we set $A=0.02$ in Equation~\ref{link_sparsity_importance} based on Ablation study~\ref{appendix:additional_exp:ablation}. We evaluate three models with three sparsity choices for each model. All the methods are aligned using Slimpajama \citep{slimpajama} as the calibration dataset, which comprises 512 sequences with a maximum length of 64 tokens.

\paragraph{Experimental Settings for Adapt-Accel and Adapt-LLMs} For Adapt-Accel and Adapt-LLMs, we keep the same choice of sparsity amplitude $A=0.02$ in Equation~\ref{link_sparsity_importance}. The global batch size is set to 128, with maximal learning rate of $2\times 10^{-5}$, and minimal learning rate of $2\times 10^{-6}$. WSD scheduler~\citep{wsd} is applied, where the combination of learning rate in all interleaved training iterations forms a WSD scheduler, with 5\% warmup steps and 10\% linear decay steps at the end. For the ShearedLLaMA~\citep{xia2024shearedllamaacceleratinglanguage} comparison, we keep these hyperparameters while allocating 10\% of tokens for pruning optimization and 90\% for post-training. Similarly, for Minitron comparison~\citep{sreenivas2024llmpruningdistillationpractice}, we keep the same settings and fine-tune the teacher model for one epoch on the dataset.

\section{Additional Experimental Results}
\label{appendix:additional_exp_results}

\subsection{Adapt-Pruner as Strong Pruners}
\label{appendix:additional_exp:pruner}

We evaluate various structured pruning methods on three LLaMA-series models: LLaMA-3.1-8B, LLaMA-3.2-3B, and LLaMA-3.2-1B~\citep{llama3modelcard}. For each model, we test sparsity levels of 20\%, 40\%, and 60\% over three trials to assess our method's effectiveness across different model sizes.

As shown in Table \ref{tbl:compare_pruning_method_details}, Adapt-Pruner demonstrates significant improvements in commonsense benchmarks, especially at 40\% sparsity level. We conjecture the sparsity of $\sim$60\% to be the ceiling of the redundancy information in LLaMA-series models, since further pruning incurs severe performance degradation, with closing gaps across different pruning methods.

\begin{table*}[h]
    \centering
    \caption{Experiment details of comparing structured pruning methods across different sparsity levels and models over three trials.} \label{tbl:compare_pruning_method_details}
    \resizebox{\linewidth}{!}{
    \begin{tabular}{lll|c|ccccccc|c|r}
        \toprule
        \toprule
        Model & Pruning Sparsity & Method & \#Param. & ARC-e & ARC-c & HellaSwag & OBQA & PIQA & SIQA & Winogrande & Average$\uparrow$ & WikiText2 ppl.$\downarrow$ \\
        \midrule
        \multirow{12}{*}{\parbox{2.4cm}{LLaMA-3.1-8B}}
         & \multirow{4}{*}{\parbox{1.8cm}{Ratio = 20\%}} & LLM-Pruner & 6.73B$\pm$0.00 & 63.92$\pm$0.76 & 39.53$\pm$1.48 & 64.85$\pm$0.99 & 38.47$\pm$1.11 & 76.64$\pm$0.47 & 42.80$\pm$0.47 & 62.85$\pm$0.81 & 55.58$\pm$0.54 & 14.78$\pm$0.25 \\
        & & FLAP & 6.48B$\pm$0.00 & 61.01$\pm$1.35 & 37.37$\pm$0.12 & 58.43$\pm$0.88 & 35.40$\pm$0.57 & 73.27$\pm$1.05 & 43.40$\pm$0.58 & 64.46$\pm$0.60 & 53.33$\pm$0.23 & 16.73$\pm$0.35 \\
        & & SliceGPT & 7.22B$\pm$0.00 & 52.27$\pm$0.33 & 29.01$\pm$0.12 & 56.37$\pm$0.58 & 32.00$\pm$0.71 & 69.84$\pm$0.56 & 41.85$\pm$0.57 & 59.46$\pm$0.55 & 48.69$\pm$0.17 & 19.61$\pm$0.14 \\
        & & Adapt-Pruner & 6.66B$\pm$0.00 & 66.43$\pm$0.85 & 39.65$\pm$0.67 & 66.71$\pm$0.49 & 38.93$\pm$0.34 & 76.87$\pm$0.16 & 44.10$\pm$0.32 & 66.19$\pm$1.41 & \bf 56.98$\pm$0.35 & \bf 14.54$\pm$0.51 \\
        \cmidrule{2-13}
         & \multirow{4}{*}{\parbox{1.8cm}{Ratio = 40\%}} & LLM-Pruner & 5.27B$\pm$0.00 & 33.90$\pm$0.48 & 21.79$\pm$0.11 & 30.00$\pm$0.29 & 25.40$\pm$1.28 & 57.02$\pm$0.08 & 35.45$\pm$0.09 & 51.30$\pm$0.73 & 36.41$\pm$0.19 & 162.81$\pm$33.99 \\
         & & FLAP & 4.98B$\pm$0.00 & 25.53$\pm$0.55 & 26.25$\pm$0.41 & 26.49$\pm$0.12 & 25.73$\pm$0.57 & 52.36$\pm$0.86 & 34.12$\pm$0.61 & 50.59$\pm$0.42 & 34.44$\pm$0.18 & 6987.58$\pm$3502.55 \\
         & & SliceGPT & 5.44B$\pm$0.00 & 33.94$\pm$0.46 & 21.79$\pm$0.44 & 33.72$\pm$0.32 & 26.27$\pm$0.41 & 57.73$\pm$0.74 & 36.44$\pm$0.25 & 49.72$\pm$0.95 & 37.09$\pm$0.20 & 85.60$\pm$1.84 \\
         & & Adapt-Pruner & 5.25B$\pm$0.00 & 45.16$\pm$0.21 & 25.97$\pm$0.26 & 44.88$\pm$0.72 & 30.40$\pm$0.85 & 66.74$\pm$0.40 & 39.03$\pm$0.24 & 56.75$\pm$0.11 & \bf 44.13$\pm$0.26 & \bf 33.75$\pm$0.25 \\
        \cmidrule{2-13}
         & \multirow{4}{*}{\parbox{1.8cm}{Ratio = 60\%}} & LLM-Pruner & 3.98B$\pm$0.00 & 26.94$\pm$1.39 & 24.43$\pm$1.65 & 26.67$\pm$0.69 & 27.93$\pm$1.59 & 51.22$\pm$0.69 & 34.14$\pm$0.26 & 49.43$\pm$1.31 & 34.39$\pm$0.07 & 2501.76$\pm$2400.07 \\
         & & FLAP & 3.71B$\pm$0.00 & 26.60$\pm$0.36 & 26.53$\pm$0.54 & 26.00$\pm$0.17 & 27.07$\pm$1.84 & 51.63$\pm$0.41 & 33.95$\pm$0.81 & 49.62$\pm$1.27 & 34.49$\pm$0.20 & 141572.73$\pm$70694.04 \\
         & & SliceGPT & 3.73B$\pm$0.00 & 28.95$\pm$0.42 & 21.61$\pm$0.53 & 28.12$\pm$0.23 & 26.13$\pm$0.66 & 53.01$\pm$0.34 & 34.49$\pm$0.33 & 49.01$\pm$0.29 & 34.48$\pm$0.07 & 218.96$\pm$2.34 \\
         & & Adapt-Pruner & 3.96B$\pm$0.00 & 32.49$\pm$0.26 & 23.64$\pm$0.43 & 30.84$\pm$0.17 & 26.40$\pm$0.49 & 56.66$\pm$0.29 & 35.59$\pm$0.36 & 49.43$\pm$1.24 & \bf 36.44$\pm$0.13 & \bf 119.95$\pm$12.88 \\
        \cmidrule{1-13}
        \cmidrule{1-13}

        \multirow{12}{*}{\parbox{2.4cm}{LLaMA-3.2-3B}}
         & \multirow{4}{*}{\parbox{1.8cm}{Ratio = 20\%}} & LLM-Pruner & 2.70B$\pm$0.00 & 56.65$\pm$2.38 & 32.14$\pm$1.33 & 55.94$\pm$0.97 & 33.60$\pm$0.28 & 73.14$\pm$1.07 & 41.79$\pm$0.65 & 56.70$\pm$0.39 & \bf 50.00$\pm$0.79 & 20.89$\pm$1.83 \\
        & & FLAP & 2.58B$\pm$0.00 & 50.42$\pm$0.70 & 29.75$\pm$0.77 & 48.87$\pm$0.83 & 31.47$\pm$0.25 & 67.99$\pm$0.53 & 41.21$\pm$0.80 & 58.04$\pm$0.14 & 46.82$\pm$0.30 & 19.90$\pm$0.26 \\
        & & SliceGPT & 3.31B$\pm$0.00 & 44.77$\pm$1.70 & 27.25$\pm$1.22 & 49.57$\pm$1.15 & 29.27$\pm$0.66 & 66.07$\pm$0.40 & 38.26$\pm$0.49 & 56.09$\pm$0.65 & 44.47$\pm$0.81 & 23.12$\pm$0.13 \\
        & & Adapt-Pruner & 2.68B$\pm$0.00 & 54.70$\pm$1.50 & 31.43$\pm$0.22 & 55.37$\pm$0.28 & 33.73$\pm$1.20 & 71.69$\pm$0.40 & 42.67$\pm$0.32 & 59.14$\pm$0.54 & \underline{49.82$\pm$0.48} & \bf 17.41$\pm$0.07 \\
        \cmidrule{2-13}
         & \multirow{4}{*}{\parbox{1.8cm}{Ratio = 40\%}} & LLM-Pruner & 2.10B$\pm$0.00 & 32.41$\pm$1.22 & 21.84$\pm$1.39 & 29.44$\pm$0.20 & 25.93$\pm$0.90 & 55.10$\pm$0.78 & 34.18$\pm$0.55 & 50.78$\pm$1.07 & 35.67$\pm$0.66 & 409.38$\pm$51.28 \\
         & & FLAP & 2.00B$\pm$0.00 & 27.06$\pm$0.59 & 24.34$\pm$1.28 & 26.15$\pm$0.25 & 27.53$\pm$1.05 & 50.36$\pm$0.44 & 35.21$\pm$0.45 & 50.49$\pm$0.49 & 34.45$\pm$0.34 & 89300.67$\pm$12876.34 \\
         & & SliceGPT & 2.52B$\pm$0.00 & 31.86$\pm$0.24 & 21.33$\pm$0.35 & 31.90$\pm$0.23 & 26.47$\pm$0.98 & 56.07$\pm$1.04 & 36.06$\pm$0.25 & 49.28$\pm$0.89 & 36.14$\pm$0.14 & 89.82$\pm$0.87 \\
         & & Adapt-Pruner & 2.06B$\pm$0.00 & 40.67$\pm$0.19 & 24.57$\pm$0.18 & 36.89$\pm$0.31 & 26.27$\pm$0.77 & 61.37$\pm$0.42 & 37.10$\pm$0.23 & 51.83$\pm$0.21 & \bf 39.82$\pm$0.24 & \bf 45.13$\pm$0.50 \\
        \cmidrule{2-13}
         & \multirow{4}{*}{\parbox{1.8cm}{Ratio = 60\%}} & LLM-Pruner & 1.59B$\pm$0.00 & 28.59$\pm$0.73 & 23.04$\pm$0.42 & 27.11$\pm$0.16 & 27.20$\pm$0.71 & 51.91$\pm$0.80 & 33.86$\pm$0.17 & 49.72$\pm$2.09 & 34.49$\pm$0.48 & 2032.20$\pm$48.33 \\
         & & FLAP &  1.50B$\pm$0.00 & 25.26$\pm$0.33 & 26.00$\pm$0.78 & 26.62$\pm$0.23 & 30.07$\pm$0.82 & 51.61$\pm$0.29 & 33.54$\pm$0.61 & 49.14$\pm$0.43 & 34.61$\pm$0.22 & 347771.12$\pm$79021.97 \\
         & & SliceGPT & 1.78B$\pm$0.00 & 28.27$\pm$0.21 & 22.12$\pm$0.21 & 28.08$\pm$0.13 & 26.13$\pm$0.50 & 52.83$\pm$0.16 & 35.38$\pm$0.02 & 50.20$\pm$0.84 & 34.72$\pm$0.06 & \bf 257.56$\pm$4.96 \\
         & & Adapt-Pruner & 1.54B$\pm$0.03 & 31.09$\pm$0.73 & 22.13$\pm$0.45 & 29.99$\pm$0.45 & 27.60$\pm$0.65 & 54.62$\pm$0.62 & 34.78$\pm$0.29 & 49.25$\pm$0.91 & \bf 35.64$\pm$0.15 & \underline{259.39$\pm$64.54} \\
        \cmidrule{1-13}
        \cmidrule{1-13}

        \multirow{12}{*}{\parbox{2.4cm}{LLaMA-3.2-1B}}
         & \multirow{4}{*}{\parbox{1.8cm}{Ratio = 20\%}} & LLM-Pruner & 1.05B$\pm$0.00 & 49.34$\pm$0.52 & 28.53$\pm$0.28 & 47.15$\pm$0.43 & 29.60$\pm$1.18 & 68.57$\pm$0.72 & 38.55$\pm$0.82 & 52.25$\pm$0.34 & 44.85$\pm$0.38 & 28.39$\pm$1.38 \\
        & & FLAP & 0.94B$\pm$0.00 & 28.73$\pm$0.09 & 24.89$\pm$0.40 & 27.38$\pm$0.15 & 26.73$\pm$0.09 & 52.03$\pm$0.72 & 34.09$\pm$0.18 & 49.12$\pm$1.19 & 34.71$\pm$0.23 & 766.81$\pm$86.10 \\
        & & SliceGPT & 1.34B$\pm$0.00 & 39.97$\pm$1.53 & 24.86$\pm$0.16 & 42.02$\pm$0.53 & 28.60$\pm$0.43 & 62.62$\pm$0.40 & 37.22$\pm$0.54 & 52.09$\pm$0.22 & 41.06$\pm$0.47 & 31.12$\pm$0.40 \\
        & & Adapt-Pruner & 1.04B$\pm$0.00 & 49.62$\pm$0.57 & 28.44$\pm$0.81 & 47.36$\pm$0.25 & 31.47$\pm$0.25 & 68.97$\pm$0.14 & 41.03$\pm$0.21 & 55.06$\pm$0.85 & \bf 45.99$\pm$0.13 & \bf 22.50$\pm$0.34 \\
        \cmidrule{2-13}
         & \multirow{4}{*}{\parbox{1.8cm}{Ratio = 40\%}} & LLM-Pruner & 0.85B$\pm$0.00 & 34.58$\pm$0.56 & 22.18$\pm$0.66 & 28.88$\pm$0.24 & 23.47$\pm$0.62 & 57.98$\pm$0.59 & 34.85$\pm$0.07 & 51.59$\pm$1.58 & 36.22$\pm$0.37 & 214.73$\pm$33.60 \\
         & & FLAP & 0.77B$\pm$0.00 & 25.87$\pm$0.62 & 26.82$\pm$0.61 & 26.80$\pm$0.08 & 27.20$\pm$0.33 & 50.87$\pm$0.19 & 34.48$\pm$0.33 & 48.75$\pm$0.64 & 34.40$\pm$0.19 & 96346.37$\pm$6993.27 \\
         & & SliceGPT & 1.06B$\pm$0.00 & 30.42$\pm$0.04 & 21.78$\pm$0.16 & 29.64$\pm$0.10 & 25.27$\pm$0.77 & 54.79$\pm$0.16 & 36.08$\pm$0.25 & 49.51$\pm$0.65 & 35.36$\pm$0.22 & 116.33$\pm$3.77 \\
         & & Adapt-Pruner & 0.84B$\pm$0.00 & 36.70$\pm$1.39 & 23.29$\pm$1.17 & 33.87$\pm$0.51 & 25.27$\pm$0.57 & 60.08$\pm$0.13 & 36.79$\pm$0.44 & 51.84$\pm$0.66 & \bf 38.26$\pm$0.44 & \bf 64.91$\pm$0.92 \\
        \cmidrule{2-13}
         & \multirow{4}{*}{\parbox{1.8cm}{Ratio = 60\%}} & LLM-Pruner & 0.67B$\pm$0.00 & 28.39$\pm$0.32 & 24.29$\pm$0.29 & 26.27$\pm$0.28 & 25.13$\pm$0.57 & 51.83$\pm$0.51 & 34.02$\pm$0.46 & 49.33$\pm$0.55 & 34.18$\pm$0.17 & 1859.01$\pm$191.57 \\
         & & FLAP & 0.62B$\pm$0.00 & 26.61$\pm$0.19 & 26.79$\pm$0.46 & 26.53$\pm$0.21 & 31.27$\pm$0.90 & 50.80$\pm$0.71 & 34.08$\pm$0.15 & 49.93$\pm$0.16 & 35.14$\pm$0.17 & 122310.98$\pm$25718.08 \\
         & & SliceGPT & 0.79B$\pm$0.00 & 28.16$\pm$0.33 & 22.98$\pm$0.18 & 27.90$\pm$0.17 & 26.33$\pm$0.25 & 53.28$\pm$0.19 & 35.19$\pm$0.30 & 48.09$\pm$0.82 & 34.56$\pm$0.14 & 380.60$\pm$9.94 \\
         & & Adapt-Pruner & 0.66B$\pm$0.00 & 32.57$\pm$0.18 & 22.35$\pm$0.86 & 29.14$\pm$0.12 & 25.00$\pm$0.33 & 54.92$\pm$0.35 & 35.28$\pm$0.15 & 50.09$\pm$1.04 & \bf 35.62$\pm$0.35 & 2\bf 95.87$\pm$31.63 \\
        
        \bottomrule
        \bottomrule
    \end{tabular}
    }
\end{table*}

\subsection{More Ablation Study}
\label{appendix:additional_exp:ablation}

\paragraph{Sensitivity Analysis of Sparsity Amplitude Parameter $A$} We empirically investigate the impact of amplitude parameter $A$ in Equation~\ref{link_sparsity_importance} on our algorithm's performance. The amplitude $A$ directly influences the architectural search space of the compressed model: insufficient amplitude constrains the exploration of potential architectural configurations, while excessive amplitude can lead to structural imbalances that degrade model performance. To systematically analyze this relationship, we conduct experiments on LLaMA-3.1-8B using a controlled setup with consistent parameters (50\% overall sparsity ratio and 50 pruning iterations) while varying the amplitude $A$ across pruning steps. This experimental design allows us to isolate and quantify the specific effects of amplitude variation on model compression outcomes.

\begin{table}[h]
    \centering
    \caption{Sensitivity analysis of amplitude parameter $A$ in progressive pruning, with 50\% sparsity and 50 pruning steps.} \label{tbl:amplitude}
    \resizebox{\linewidth}{!}{
    \begin{tabular}{l|c|ccccccc|c|r}
        \toprule
        \toprule
        Amplitude $A$ & Para. Num. & ARC-e & ARC-c & BoolQ & HellaSwag & OBQA & PIQA & Winogrande & Average$\uparrow$ & WikiText2 ppl.$\downarrow$ \\
        \midrule
        0 & 4.54B & 29.04 & 21.84 & 45.47 & 28.22 & 27.00 & 53.81 & 48.30 & 36.24 & 621.03 \\
        0.005 & 4.62B & 37.25 & 22.87 & 55.90 & 33.89 & 27.80 & 59.52 & 52.64 & 41.41 & 122.04 \\
        0.01 & 4.61B & 37.29 & 25.17 & 60.73 & 35.60 & 28.80 & 60.77 & 52.09 & 42.92 & 95.59 \\
        0.02 & 4.43B & 40.61 & 26.45 & 62.05 & 35.18 & 28.20 & 62.30 & 53.67 & \bf 44.07 & \bf 90.80 \\
        0.04 & 4.04B & 34.51 & 24.06 & 56.21 & 31.17 & 26.40 & 57.34 & 51.78 & 40.21 & 314.90 \\
        \bottomrule
        \bottomrule
    \end{tabular}
    }
\end{table} 

As demonstrated in Table \ref{tbl:amplitude}, the baseline case of $A=0$ represents uniform pruning without our proposed adaptive mechanism. Notably, varying the amplitude $A$ yields different final parameter counts due to its influence on architectural decisions during compression. Our experimental results reveal that $A=0.02$ achieves optimal performance, maximizing the benchmark average while minimizing model complexity on WikiText2. The performance exhibits a non-monotonic relationship with $A$: as amplitude increases, model performance initially improves before degrading, consistent with the theoretical trade-off between exploration capacity and architectural stability.

\end{document}